\documentclass[letterpaper]{article}
\usepackage{aaai24}
\usepackage{times}
\usepackage{helvet}
\usepackage{courier}
\usepackage[hyphens]{url} 
\usepackage{graphicx}
\usepackage{natbib}
\usepackage{caption}
\usepackage{algorithm}
\usepackage{algorithmic}
\usepackage{subcaption}
\usepackage{booktabs}
\usepackage{multirow}
\usepackage{tabularx} 
\usepackage{comment}
\usepackage{xspace}
\usepackage{makecell}
\usepackage{mathtools}
\usepackage{enumitem}
\usepackage{tikz}
\usepackage{textcomp}
\usepackage{amsmath,amsfonts}
\usepackage[noabbrev,capitalize]{cleveref}
\usepackage{url}
\usepackage{newfloat}
\usepackage{listings}
\usepackage{bibentry}
\usepackage{dsfont}

\DeclareCaptionStyle{ruled}{labelfont=normalfont,labelsep=colon,strut=off}
\floatstyle{ruled}
\newfloat{listing}{tb}{lst}{}
\floatname{listing}{Listing}
\title{Benchmarking Evolutionary Community Detection Algorithms\\in Dynamic Networks}
\author{
    Giordano Paoletti\textsuperscript{\rm 1},
    Luca Gioacchini\textsuperscript{\rm 1},
    Marco Mellia\textsuperscript{\rm 1},
    Luca Vassio\textsuperscript{\rm 1},
    Jussara M. Almeida\textsuperscript{\rm 2}
}
\affiliations{
    \textsuperscript{\rm 1}Politecnico di Torino, Turin, Italy\\
    first.last@polito.it\\
    \textsuperscript{\rm 2}Universidade Federal de Minas Gerais, Belo Horizonte, Brazil\\
    jussara@dcc.ufmg.br
}

\begin{document}

\maketitle

\pubnote{Accepted at the 4\textsuperscript{th} Workshop on Graphs and more Complex structures for Learning and Reasoning (GCLR) at AAAI 2024}

\newcommand{\ie}{\mbox{i.e.}\xspace}
\newcommand{\eg}{\mbox{e.g.}\xspace}

\begin{abstract}

In dynamic complex networks, entities interact and form network communities that evolve over time. Among the many \emph{static} Community Detection (CD) solutions, the modularity-based Louvain, or Greedy Modularity Algorithm (GMA), is widely employed in real-world applications due to its intuitiveness and scalability. Nevertheless, addressing CD in dynamic graphs remains an open problem, since the evolution of the network connections may poison the identification of communities, which may be evolving at a slower pace. Hence, na\"ively applying GMA to successive network snapshots may lead to temporal inconsistencies in the communities. Two evolutionary adaptations of GMA, sGMA and $\alpha$GMA, have been proposed to tackle this problem. Yet, evaluating the performance of these methods and understanding to which scenarios each one is better suited is challenging because of the lack of a comprehensive set of metrics and a consistent ground truth. To address these challenges, we propose (i) a benchmarking framework for evolutionary CD algorithms in dynamic networks and (ii) a generalised modularity-based approach (NeGMA). Our framework allows us to generate synthetic community-structured graphs and design evolving scenarios with nine basic graph transformations occurring at different rates. We evaluate performance through three metrics we define, \ie Correctness, Delay, and Stability. Our findings reveal that $\alpha$GMA is well-suited for detecting intermittent transformations, but struggles with abrupt changes; sGMA achieves superior stability, but fails to detect emerging communities; and NeGMA appears a well-balanced solution, excelling in responsiveness and instantaneous transformations detection.

\end{abstract}

\section{Introduction}
\label{s:introduction}

Dynamic complex networks, or graphs, serve as a powerful tool to model computer, social or generic communications where entities (nodes) interact with each other establishing connections (edges) that evolve over time~\cite{wang2019time,dakiche2019survey}. Detecting communities, or groups of nodes that exhibit stronger internal connections compared to their connections with the broader network~\cite{girvan2002community}, and understanding their temporal dynamics is crucial for a deeper comprehension of the underlying structure and function of the systems they represent.

The scientific literature is rich in Community Detection (CD) approaches for static graphs, like the application of spectral clustering algorithms to the graph Laplacian~\cite{ng2001spectral,dall2020spectral2}, hierarchical CD~\cite{li2022hierarchical} or statistical modelling~\cite{peixoto2019statistical,geng2019probabilistic}, just to cite few. Among them, the modularity-based Louvain, or Greedy Modularity Algorithm (GMA)~\cite{blondel2008louvain}, is one of the most used methods for real-world applications. It is a fast and scalable solution to detect an unspecified number of communities in both small and large heterogeneous networks by optimising their \emph{modularity}~\cite{newman2006modularity}, an intuitive and easily understandable measure of a graph tendency to form communities.

Nevertheless, due to the ever-changing nature of connections, network topologies are constantly changing. Such changes may or may not represent a natural evolution of the underlying communities. 
For example, short-term changes in node and edge sets might suggest substantial changes. However, beneath these rapid fluctuations, the underlying communities might remain consistent or evolve at a significantly slower pace than the observed changes suggest. Finding effective solutions to address CD problems in dynamic graphs is still an arduous challenge~\cite{dakiche2019survey}. A na\"ive approach involves independently running GMA on successive snapshots of the dynamic graph. Nevertheless, this necessitates an additional step to align communities across time, typically by comparing membership overlap~\cite{spiliopoulou2013monic}, which can result in a loss of temporal coherence.
To our knowledge, the first attempt to adapt GMA for evolutionary scenarios is the Stabilised Louvain (sGMA)~\cite{aynaud2010smoothed}. This method iteratively applies the standard Louvain over snapshots by initialising nodes as members of their previously assigned communities. Another approach is Louvain with memory ($\alpha$GMA)~\cite{elegazzar2021alpha}. This approach incorporates a memory term into the weights of the graph edges prior to the application of standard GMA.

\begin{figure}[!t]
    \centering
    \begin{subfigure}[b]{.5\linewidth}
        \centering
        \includegraphics[width=\linewidth]{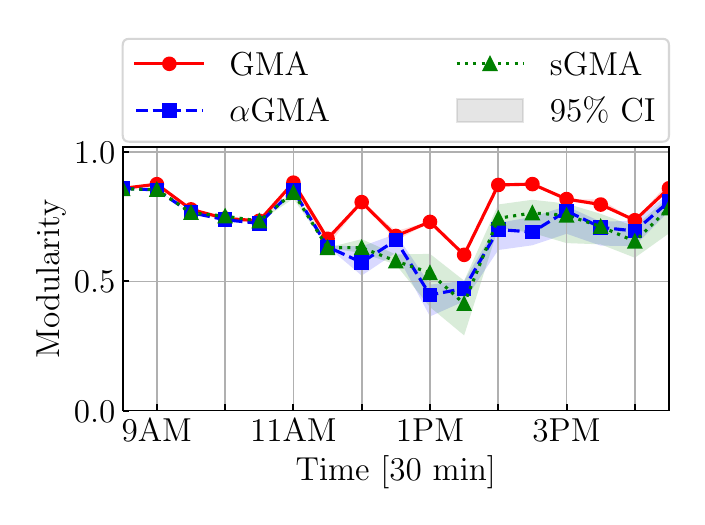}
        \caption{Modularity.}
        \label{fig:real_modularity}
    \end{subfigure}
    \hspace{-.5em}
    \begin{subfigure}[b]{.5\linewidth}
        \centering
        \includegraphics[width=\linewidth]{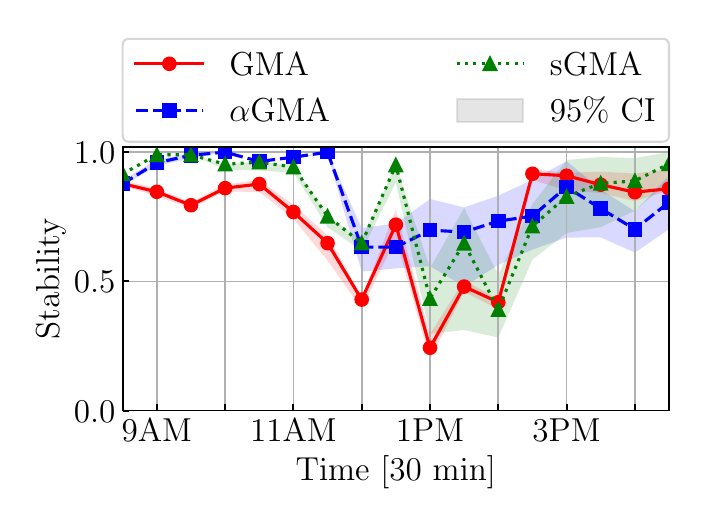}
        \caption{Algorithms stability.}
        \label{fig:real_stability}
    \end{subfigure}
    \caption{Evaluation of static and evolutionary CD algorithms on the real-world \texttt{SocioPatterns} dataset.}
    \label{fig:realcase}
\end{figure}

While these algorithms successfully maintain temporal consistency in community assignments, their relative performance in different scenarios of community evolution is still unclear. For illustration purposes, in \cref{fig:realcase} we consider the detection of communities in the  \texttt{SocioPatterns} dataset~\cite{stehle2011dataset}, which records the interactions between children and teachers in ten classes of a primary school during a day. To that end, we run 50 instances of independent GMA as well as the two aforementioned evolutionary extensions, comparing their performance with respect to graph modularity and stability -- \ie how similar the output is to the previous snapshot, measured every 30 minutes (see Section \ref{ss:metrics}). The latter aims at capturing the consistency of communities over time, which is a desirable property in a realistic evolving scenario.

We observe two major effects: for stability, GMA tends to greatly change the communities across consecutive snapshots due to the natural randomness of contacts -- \ie in many snapshots, stability drops significantly, and both $\alpha$GMA and sGMA improve it.
Yet, GMA tends to offer better performance with respect to modularity, since it provides an independent and unconstrained solution for each snapshot. 
These results highlight the need for a trade-off between different community quality metrics in dynamic environments. Different CD approaches address such a trade-off in different ways. 
As such, it calls for a deeper study of the (relative) performance of evolutionary CD algorithms under different and controllable scenarios.

In this paper, we address these challenges by proposing a framework for benchmarking evolutionary modularity-based CD algorithms in dynamic networks. To evaluate alternative methods with respect to known ground truth, we generate community-structured graphs through the Lancichinetti-Fortunato-Radicchi (LFR)~\cite{lancichinetti2008benchmark} benchmark and design 9 \emph{in-vitro} graph transformations, reflecting evolutionary patterns that may arise in real-world scenarios and occurring at different rates.  
We evaluate the performance of evolutionary GMA solutions defining three metrics that capture complementary and yet desirable properties. Namely, \emph{Correctness} and \emph{Delay} to detect the transformations, which are supervised metrics leveraging the generated ground truth, and the aforementioned unsupervised \emph{Stability} metric. To the best of our knowledge, we are the first proposing metrics specifically tailored for this goal.

Furthermore, we introduce a generalised evolutionary algorithm called Neighbourhood-based Louvain (NeGMA) and compare it against the existing $\alpha$GMA and sGMA through the designed framework.

Our findings reveal that (i) $\alpha$GMA achieves high stability in intermittent transformations, failing in detecting instantaneous transformations; (ii) sGMA outperforms all the tested solutions in terms of temporal coherence, but prevents the detection of new emerging communities; (iii) NeGMA is a well-balanced CD algorithm overcoming the limitation of existing solutions and proving high responsiveness and detection rates for instantaneous transformations.

\section{Framework Overview}
\label{s:framework}

We define a dynamic graph $\{ \mathcal{G}^t \}_{t=1}^N$ as a sequence of $N$ undirected graphs $\mathcal{G}^t=(\mathcal{V}^t,\mathcal{E}^t)$, where $\mathcal{V}^t$ and $\mathcal{E}^t$ are, respectively, the set of nodes and edges at snapshot $t$\footnote{Possibly, each snapshot can represent a single edge or node change.}. 
Namely, an edge $\epsilon = (u,v,w^t)\in \mathcal{E}^t$ indicates a connection between nodes $u,v \in \mathcal{V}^t$ with weight $w^t \in \mathbb (0,1]$ at snapshot $t$.
Focusing on a single snapshot\footnote{We omit the suffix $t$ for the sake of clarity.}, we define a community $c \subseteq \mathcal{V}$ as a set of nodes with a higher density of internal connections compared to their connections with the rest of the network. 

Given a set of non-overlapping communities $C$ such that $\bigcup_{c \in C} c = \mathcal{V}$, we rely on the graph \emph{modularity}~\cite{newman2006modularity} to quantify how well-defined the communities are. In a nutshell, modularity compares the density of the within-community connections to the connections expected in a random network.
Formally, for community $c\in C$, its modularity is defined as $Q_c = \frac{w_{in, c}}{w^*} - \left(\frac{w_c}{2w^*}\right)^2$, where $w^*$ is the sum of all edge weights of the graph, $w_{in, c}$ is the sum of the intra-community edge weights, and $w_c$ is the sum of the weighted degree of the nodes in community $c$. The modularity of the graph is defined as the sum of the modularity of its communities $Q = \sum_{c \in C} Q_c$.

GMA~\cite{blondel2008louvain} is designed to (greedily) maximise modularity.

\subsection{Generating synthetic graphs}

The evaluation of CD algorithms is often challenged by the lack of ground truth. It becomes even harder in dynamic scenarios when, as illustrated in \cref{fig:realcase}, metrics capturing different desired properties of a solution may produce quite opposite results. Thus, to conduct a sound and thorough evaluation of alternative evolutionary CD methods, we argue for the use of synthetic scenarios reflecting transformations in the graph topology that occur in real-world evolving networks and offer a known ground truth against which alternative solutions can be compared.

Specifically, we run \emph{in-vitro}  experiments generating synthetic graphs through the Lancichinetti-Fortunato-Radicchi (LFR) benchmark~\cite{lancichinetti2008benchmark}, which is a widely recognised and versatile tool for creating complex networks exhibiting community structures and providing known ground-truth (\ie the assigned communities). Any synthetic community-based graph generation algorithms can be used. LFR graphs are characterised by power-law degree distribution and heterogeneous community sizes. We randomly assign weights $\in (0,1]$.
\begin{table}[!t]
\centering
\footnotesize
\setlength{\tabcolsep}{2pt}
\caption{Overview of scenarios and transformations.}
\label{tab:events}
\begin{tabular}{lll}
\toprule
\textbf{Scenario} & \textbf{Transf.} & \textbf{Description} \\
\midrule
\multirow{3}{*}{Noise} & Expansion & Add nodes within communities\\
 & Intermittence & Turn off/on nodes for \emph{one} snapshot \\
 & Switch & Turn off/on nodes for \emph{multiple} snapshots \\
\midrule
\multirow{4}{*}{Morphing} & Merge & Merge community pairs \\
 & Split & Form sub-communities \\
 & Death & Dissolve communities \\
 & Birth & Aggregate nodes in new communities \\
\midrule
\multirow{2}{*}{Disruptive} & Mixing & Decay of community structure \\
 & Removal & Remove nodes \\
 \bottomrule
\end{tabular}
\end{table}

\subsection{Simulated transformations and scenarios}
\label{ss:scenarios}
Our framework creates 9 transformations, grouped into 
 3 scenarios -- \ie (i) Noise, (ii) Morphing and (iii) Disruptive. These transformations can be applied at different rates. We provide a concise overview of the designed scenarios in \cref{tab:events}.

\paragraph*{Noise} In this scenario we evaluate the CD algorithm's robustness and stability against perturbations over time. 
We perturb the graph without affecting the initial ground truth provided by the LFR graph through 3 transformations: (i) \emph{Expansion}. At each snapshot, we progressively add new nodes to existing communities while preserving node degree heterogeneity through preferential attachment~\cite{Baraba_si_1999}. The new nodes persist throughout the simulation; (ii) \emph{Intermittence}.
At each snapshot $t$, we remove a fraction of nodes $\phi_{int}$ and we re-add them at $t+1$. To ensure the persistence of hub nodes characterising the communities, each node $k$ has probability $p_{k} \propto \frac{1}{\delta_k}$ of being selected, where $\delta_k$ is the node degree; and (iii) \emph{Switch}. At each snapshot $t$, we remove a fraction of nodes $\phi_{swi}$. Each node $k$ is removed with probability $p_{k}$, as in the intermittence case. Inactive nodes reappear with probability $p_{swi} \propto \exp^{t_{off} -\gamma }$, where $\gamma \in \mathbb{N}$ is a user-defined parameter and $t_{off}$ is the number of snapshots for which the node has been removed.

These transformations reflect real-world scenarios, such as user behaviour in a social media platform, where user activity exhibits an on-off pattern and new users may join the system without necessarily changing the underlying community structure.

\paragraph*{Morphing} In this scenario we progressively change the community structure and the network topology over time. 
We start from the initial ground truth GT$^0$ for communities provided by the LFR and, after $N$ snapshots, we end up with a different ground truth GT$^N$ (hence different communities) which we define according to the considered transformation.
We design 4 transformations: (i) \emph{Split}. We select one or more communities and progressively split them into two sub-communities by loosening their edge weights; {(ii) \emph{Merge}}. We select one or more community pairs and progressively merge them into a larger one by tightening the edge weights; (iii) \emph{Death}. We select one (or more) communities and progressively dissolve them by loosening the intra-community edge weights and linking the nodes to other existing communities through preferential attachment; (iv) \emph{Birth}. We progressively loosen the edge weights of some nodes per community and link them together forming new communities. We control the speed at which the transformations occur through the parameter $\tau \in[0, 1]$, which expresses the variation in edge weights at each snapshot. At snapshot $t$ we obtain the weight of an edge $w^t$ by modifying the weight of the previous snapshot $w^{t-1}$. Formally, $w^t = \tau \pm w^{t-1}$. Notice that we clip $w^t \in (0, 1]$.

These transformations reflect transitions in real-world networks, such as the rise of new communities in social networks centred on a specific topic or, conversely, the waning interest in certain subjects. 

\paragraph*{Disruptive} 
In this scenario, we progressively destroy the community structure of the network. We define two transformations: (i) \emph{Mixing}. At each snapshot, we shuffle the initial communities replacing the destination node of some edges with different existing nodes, preserving the weighted degree distribution; (ii) \emph{Removal}. At each snapshot, we progressively remove a fraction of nodes $\phi_{rem}$ selected uniformly at random.

These transformations reflect real-world scenarios marked by a shift towards a disordered arrangement of nodes, like social mixing, where nodes from different communities randomly connect with each other.

\subsection{Quality metrics}
\label{ss:metrics}

While modularity is the gold standard for evaluating community assignment, \cref{fig:realcase} highlights that one single metric does not suffice for a complete evaluation of evolutionary cases. Thus, we define 3 metrics which capture complementary factors: \emph{Stability} (unsupervised), \emph{Correctness} and \emph{Delay} (both use the information of the ground truth, hence supervised). These metrics rely on the Adjusted Mutual Information (AMI)~\cite{vinh2009ami}, which measures the level of mutual information shared between two sets of communities accounting for unbalanced community sizes.

\paragraph*{Stability ($S$)}  Formally, {$S^t=\text{AMI}(C^{t-1},C^t)$}. It is a proxy for the temporal consistency in community assignment. We obtain the overall stability $S$ by averaging $S^t$ over the simulation snapshots. A higher stability indicates fewer changes in community membership between two snapshots.

\paragraph*{Correctness ($K$)} Given a ground truth (GT), we evaluate the correctness of the considered algorithm in detecting the transformations at snapshot $t$ as $K^t = \text{AMI}(\text{GT},C^t)$.

In the noise scenario, where the transformations do not affect the initial ground truth, we set GT=GT$^0$. Additionally, we obtain the overall correctness $K$ averaging $K^t$ over the simulated snapshots.

Conversely, in the morphing scenario, the final ground truth is different from the initial one. Thus, we set GT=GT$^N$. Since we cannot define the final ground truth until the transformation has ended, we consider the overall correctness $K$ as $K^t$ at the final snapshot. Formally, $K=K^N$.

In disruptive transformations, the final ground truth GT$^N$ is not available and we do not consider the correctness in this scenario.

\paragraph*{Delay ($D$)} It is the number of snapshots the algorithm takes to detect the change to a new community structure. As such, we define this metric only in morphing transformations and use it to assess the \emph{responsiveness} of each algorithm to detect the transformation. Namely, we define the Crossing Point CP as the first snapshot at which the community assignment is closer to GT$^N$ compared to GT$^0$. Formally, $\text{CP}=\min\{t:\text{AMI}(\text{GT}^N,C^t)>\text{AMI}(\text{GT}^0,C^t)\}$ Hence, we consider the delay $D$ as the difference between the CP and the snapshot at which the transformation begins. If CP does not exist, we set the delay to the maximum possible value of $t$.

\begin{table*}[]
\centering
\footnotesize
\caption{Overview of Correctness $K$ and Stability $S$ of the tested evolutionary algorithms compared to the median of the independent GMA (baseline). `=' indicates a performance
variation $\leq 0.5\%$. A number $n$ of `+' indicates an improvement in the $(\;5(n-1)\%\;,5n\%\;]$ range. A number $n$ of `--' indicates a decrease in the $[\;-5 n\%\;,-5(n-1)\%\;)$ range.}
\label{tab:main_results}
\begin{tabular}{ll|ccc|cccc|cc}
\toprule
& & \multicolumn{3}{c|}{\textbf{Noise Scenario}} & \multicolumn{4}{c|}{\textbf{Morphing Scenario}} & \multicolumn{2}{c}{\textbf{Disruptive Scenario}} \\
&\textbf{} & \textbf{Expansion} & \textbf{Interm.} & \textbf{Switch} & \textbf{Merge} & \textbf{Split} & \textbf{Birth} & \textbf{Death} & \textbf{Mixing} & \textbf{Removal} \\
\midrule
\multirow{4}{*}{$K ^\dagger$} &
$\alpha$GMA  & = & + & =   & = & =     & =     & =   & n.a. & n.a. \\
& sGMA       & = & -- -- & =    & = & -- -- -- -- & -- -- -- -- -- & =   & n.a. & n.a. \\
& NeGMA      & = & -- & =    & = & -- --     & -- --    & =   & n.a. & n.a. \\
\cmidrule{2-11}
& GMA      & 0.84 & 0.95 &  0.99 & 1.00 & 0.86 & 0.91 & 1.00  & n.a. & n.a. \\
\midrule
\multirow{4}{*}{$S$} & 
$\alpha$GMA  & = & +++ & =  &= & =     & --   & =    & = & = \\
& sGMA       & + & ++ & =   &+ & ++    & +++ & +++  & = & = \\
& NeGMA      & = & + & =    &+ & +    & ++  & ++   & = & = \\
\cmidrule{2-11}
& GMA      & 0.98 & 0.90 &  1.00 & 0.98 & 0.93 &  0.88 & 0.87 & 0.99 & 0.99 \\
\bottomrule
\multicolumn{11}{l}{$^\dagger$ Recall that we compute differently the correctness for noise and morphing scenarios (see \cref{ss:metrics}).} \\
\end{tabular}
\end{table*}

\section{Modularity-based CD}
\label{s:algorithms}

The Louvain method, also known as the Greedy Modularity Algorithm (GMA)~\cite{blondel2008louvain}, is a widely used and efficient approach for CD in static networks. It operates by iteratively greedily optimising the graph modularity. In a nutshell, at the first iteration, GMA assigns each node at random to a unique community. At each subsequent iteration, it merges communities by evaluating the modularity gain resulting from moving a node to a neighbouring community or merging two adjacent communities. If the modularity gain is positive, the change is accepted, and the process continues. We invite the reader to refer to ~\cite{blondel2008louvain} for details.

In the following, we consider evolutionary CD algorithms that use GMA as a starting point. A na\"ive evolutionary extension consists of applying GMA on each snapshot independently. We refer to this approach as \emph{independent GMA}. 

\subsection{Louvain with memory ($\alpha$GMA)} 
This approach~\cite{elegazzar2021alpha} consists of prioritising the persistence of certain edges over time, favouring past partitions over new points to maintain community stability.  In a nutshell, $\alpha$GMA introduces a memory term  $\alpha$ in the graph definition (notably in the edge weights) updating the network weights at snapshot $t$ according to its history at snapshot $t-1$.  
Namely, an edge $\epsilon$ whose weight during snapshot $t$ is  $\hat{w}^t$ has its actual weight updated to
{$w^t = (1-\alpha)\cdot \hat{w}^t + \alpha \cdot  w^{t-1}$} if $w^{t-1}\neq 0$. Otherwise, if $w^{t-1} = 0$, we set $w^t = \hat{w}^t$.   
The application of independent GMA on such a graph results in gradual modularity adjustments as new information becomes available.

\subsection{Stabilised Louvain (sGMA)} This algorithm changes the initialisation process of GMA. Unlike the traditional method, which assigns each node to a unique distinct community, at snapshot $t$ sGMA maintains the previous community membership for nodes active also in $t-1$. Each new node is assigned to a new distinct community. Given this initialisation, we run the GMA allowing nodes to adjust their community membership if the initial assignment is not suitable.

\subsection{Novel Neighbourhood-based GMA (NeGMA)}

\begin{figure}[!t]
    \centering
    \includegraphics[width=\linewidth]{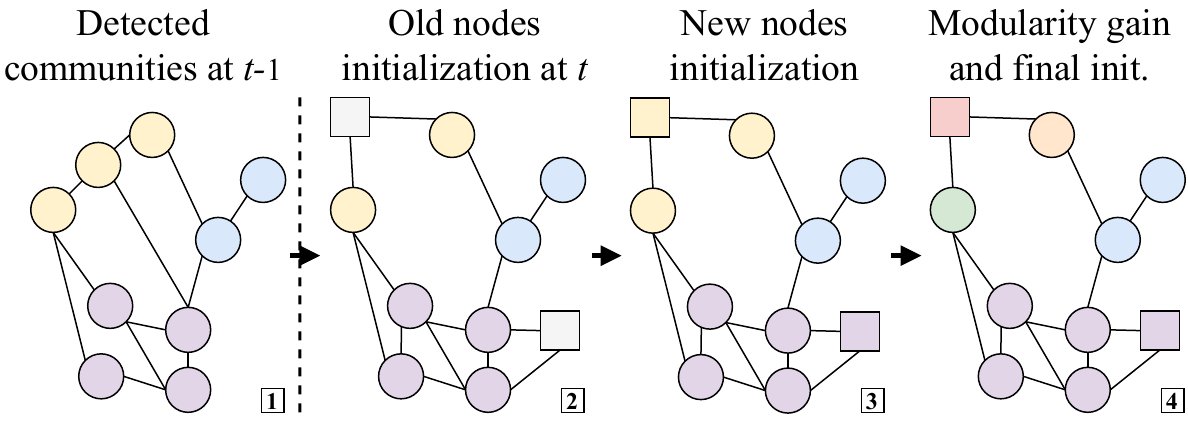}
    \caption{Proposed NeGMA initialisation of nodes at snapshot $t$ starting from snapshot $t-1$. Different colours represent different communities; squares represent new nodes.}
    \label{fig:negma}
\end{figure}

Both $\alpha$GMA and sGMA employ strategies to maintain stability in community assignment by keeping the new community structure as similar as possible to the previous snapshot. By doing so, both methods are expected to be better suited for detecting few or slow-rated modifications. Yet, they may struggle to properly track scenarios of greater or more abrupt changes. 
We here address this issue by (i) modifying the sGMA initialisation of new nodes which were inactive in the network relying on their neighbourhood in the current snapshot and (ii) evaluating the local community modularity $Q_c$ after this novel initialisation.
We call our approach NeGMA and provide an overview in \cref{fig:negma}. 

Namely, (i) we start from the communities detected at the previous snapshot $t-1$. At snapshot $t$ the graph evolves, some new edges and nodes appear, and some disappear; (ii) for already existing nodes we assign them to the communities detected at $t-1$ (as in sGMA); (iii) for new nodes, we assign the community through majority voting on the communities of their \emph{neighbourhood}; (iv) for each community, we compute the local modularity gain with respect to the ones of the previous snapshot: $\Delta Q_c^t = Q_c^{t} - Q_c^{t-1}$. If $\Delta Q_c^t$ is lower than a user-defined threshold $\theta_Q$, we unbind the community assigning each of its nodes to a distinct community like traditional GMA (former yellow community in the upper-left part of the graph at step 4 in \cref{fig:negma}).

\section{Stability and Correctness Evaluation}
\label{s:experiments}

We start our evaluation by assessing the stability and correctness of the CD algorithms under the various transformations presented in \cref{tab:events}, deferring the discussion on delay to the next section.
We run our experiments generating 100 independent graphs for each transformation. We set the LFR parameter $\mu=0.2$~\cite{lancichinetti2008benchmark} and we generate $N=150$ temporal snapshots. We trigger the beginning of the transformations at snapshot 25 and conclude it at snapshot 125. For each temporal graph, we perform 10 independent runs of each algorithm on each graph.
To address the variability introduced by (i) the randomness of the algorithms and (ii) the graph generation, we initially average the results of the 10 runs on each graph. Consequently, we evaluate the median and bootstrapped 99\% confidence interval for these medians across the 100 graphs. 

Regarding the transformations, we set the fraction of nodes for intermittence as $\phi_{int}=0.2$; for switch $\phi_{swi}=0.005$ and $\gamma=10$; for removal, we uniformly sample $\phi_{rem}$ between $[0.005, 0.02]$. When not explicitly stated, we set the transformation speed $\tau =$ 0.01. 
As suggested by authors of $\alpha$GMA~\cite{elegazzar2021alpha}, we set $\alpha=0.8$, whereas for NeGMA we set $\theta_Q = 0$\footnote{Notice that in case of modularity decrease for all the communities, NeGMA corresponds to independent GMA. We plan to evaluate the impact of $\theta_Q$ in future developments.}.
Finally, we run our experiments on a commodity server with 72 CPUs. We hope our framework and results can inspire other works towards the analysis of evolutionary CD. For that, we release our source code and generated graphs upon request.

In \cref{tab:main_results} we provide a bird's eye overview of the algorithms' performance across all simulated scenarios. The signs refer to the relative performance gain/loss obtained by each algorithm over using independent GMA, considered as baseline. For reference, we report also the values of the baseline.

Firstly, independent GMA achieves satisfactory performance in noisy scenarios (overall correctness $>80\%$ and stability $\geq90\%$). However, its stability is notably impacted by morphing transformations, dropping to  87\% and 88\% in birth and death transformations, respectively. 

By preserving existing connections across successive snapshots, $\alpha$GMA enhances correctness and, notably, stability under intermittence noise (both peaking at $\approx 100\%$), but delivers similar performance in all other transformations (with a small loss in stability for birth). 

Both sGMA and NeGMA offer notable improvements in stability in scenarios when $\alpha$GMA fails to do so -- \ie morphing transformations. Nevertheless, such improvements come at the expense of losses in correctness, especially for transformations leading to the emergence of new communities (\ie split and birth). Yet, compared to sGMA, NeGMA offers a better trade-off between stability and correctness in such scenarios. 

Finally, all the algorithms demonstrate remarkable stability (99\%) performing on par with the baseline.

\begin{figure}[!t]
    \centering
    \begin{subfigure}[b]{.5\linewidth}
        \centering
        \includegraphics[width=\linewidth]{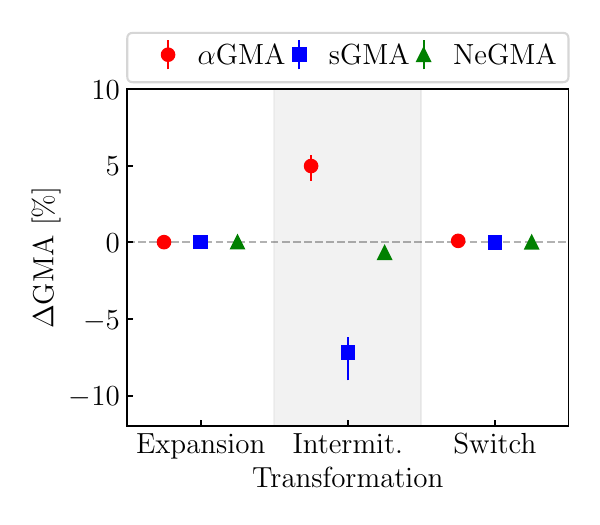}
        \caption{Correctness}
        \label{fig:noise_correctness}
    \end{subfigure}
    \hspace{-.5em}
    \begin{subfigure}[b]{.5\linewidth}
        \centering
        \includegraphics[width=\linewidth]{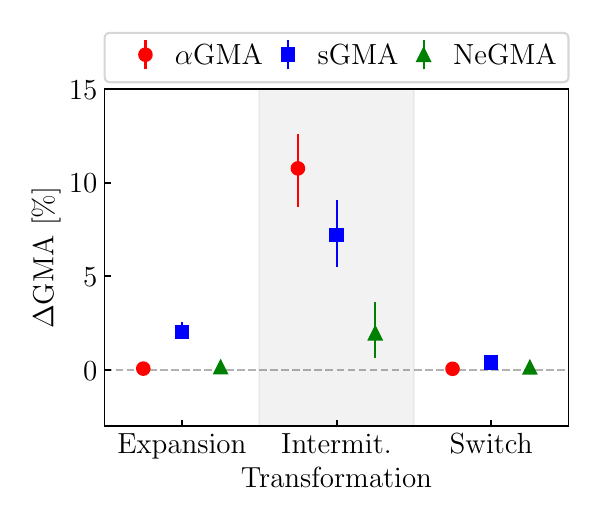}
        \caption{Stability}
        \label{fig:noise_stability}
    \end{subfigure}
    \caption{\emph{Noise}. Median gain and bootstrapped 99\% confidence interval in Correctness and Stability.}
    \label{fig:noise}
\end{figure}

\begin{figure*}[!t]
    \centering
    \begin{subfigure}[b]{.24\linewidth}
        \centering
        \includegraphics[width=\linewidth]{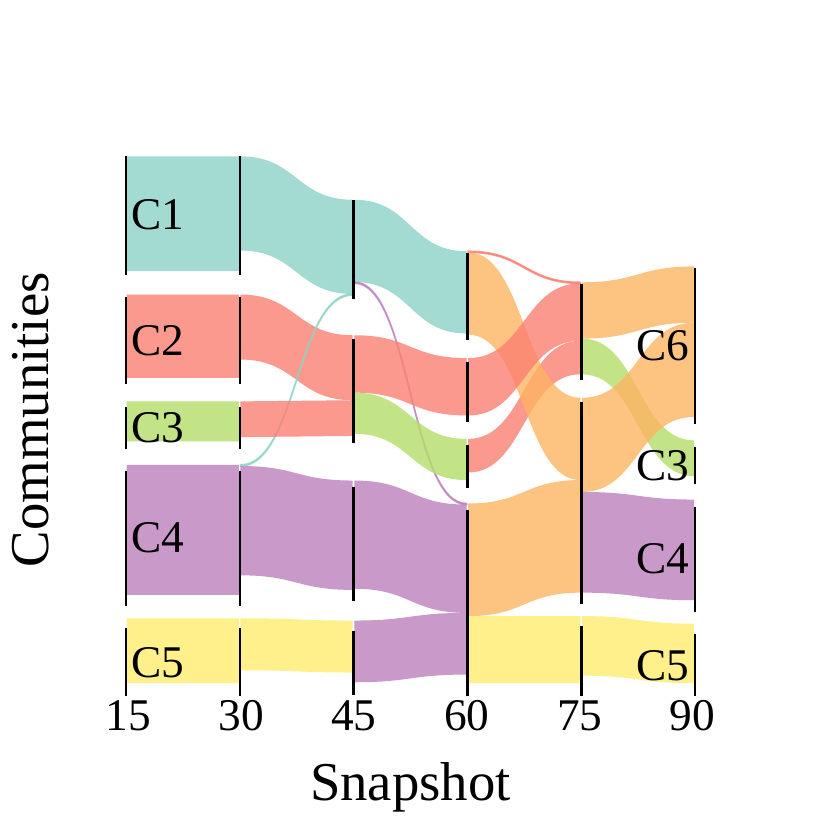}
        \caption{GMA}
        \label{fig:sankey_noise_GMA}
    \end{subfigure}
    \begin{subfigure}[b]{.24\linewidth}
        \centering
        \includegraphics[width=\linewidth]{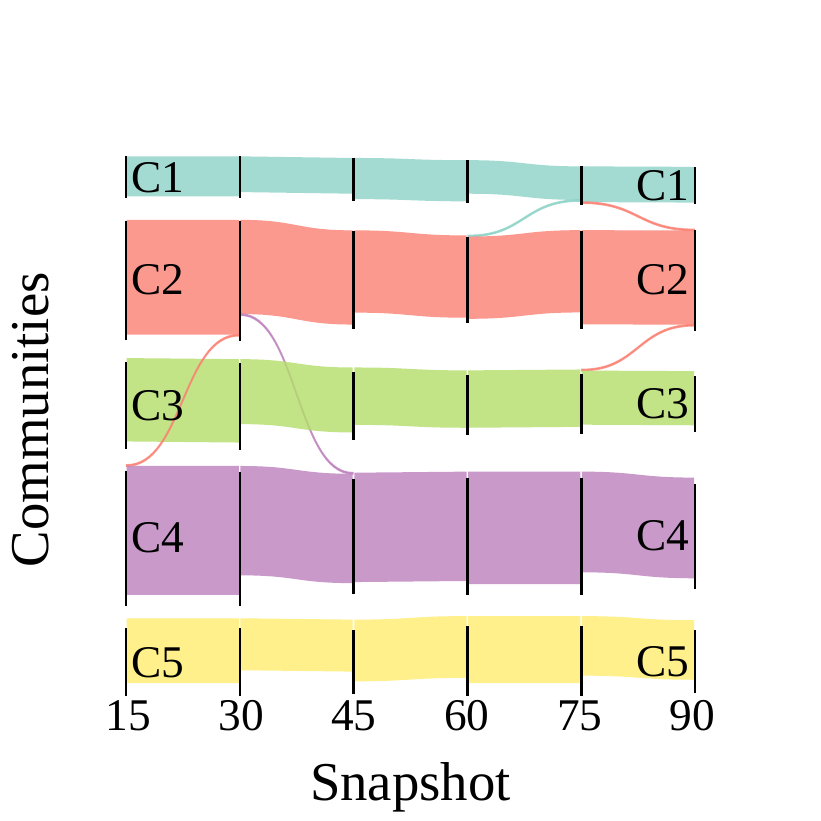}
        \caption{$\alpha$GMA}
        \label{fig:sankey_noise_aGMA}
    \end{subfigure}
        \begin{subfigure}[b]{.24\linewidth}
        \centering
        \includegraphics[width=\linewidth]{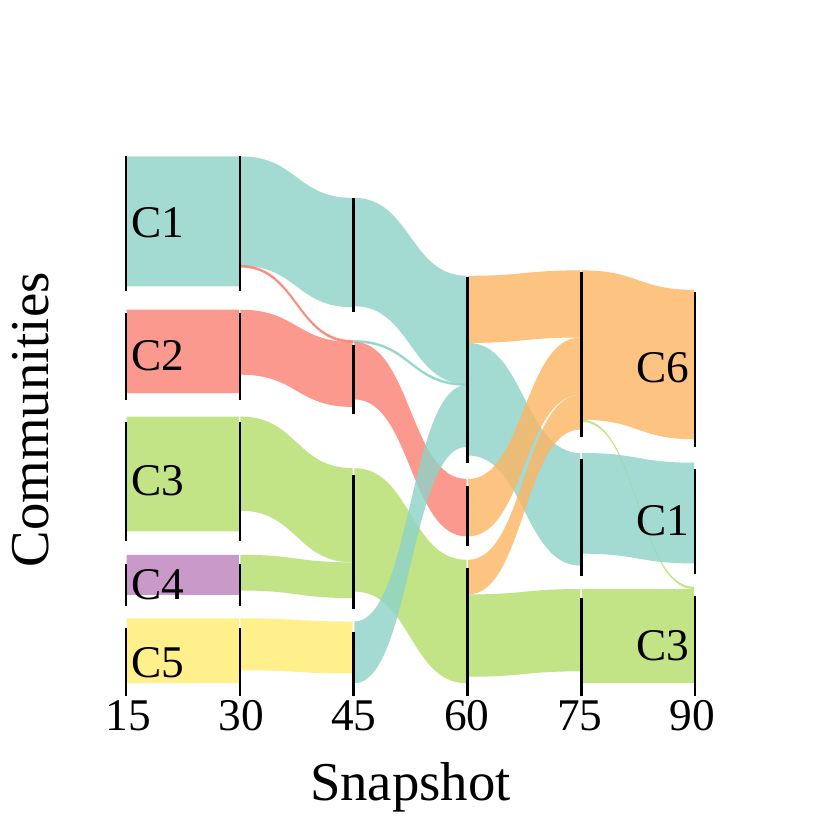}
        \caption{sGMA}
        \label{fig:sankey_noise_sGMA}
    \end{subfigure}
    \begin{subfigure}[b]{.24\linewidth}
        \centering
        \includegraphics[width=\linewidth]{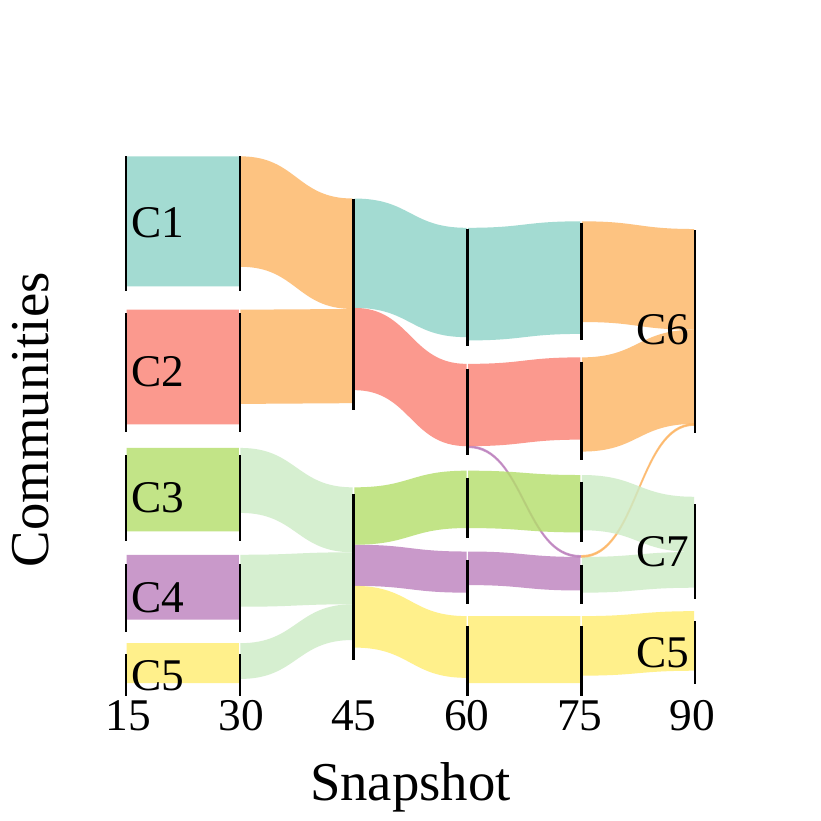}
        \caption{NeGMA}
        \label{fig:sankey_noise_neGMA}
    \end{subfigure}
    \caption{Dynamic evolution of the graph perturbed by the intermittence transformation, showcasing community assignment. Zoom in a sample of snapshots from 15 to 90 with step 15 where changes occur.}
    \label{fig:sankey_intermittence}
\end{figure*} 
As argued, there is no single winner across all scenarios, and each algorithm offers a different trade-off between correctness and stability, depending on the transformation. Next, we delve deeper into these results and expand our discussion on how each algorithm behaves for these transformations.

\paragraph*{Noise transformations}
In \cref{fig:noise} we report {\it median} correctness and stability gain/loss compared to independent GMA in noise scenario, spanning from snapshots 25 to 125.   
All tested algorithms perform on par with GMA for the expansion and switch transformations, yielding a maximum gain of 2\%.

The memory term integrated into $\alpha$GMA graph generation smooths the effects of the transformation, enhancing robustness against intermittent noise by boosting both correctness ($+5\%$) and stability ($+11\%$).
This can be observed in the qualitative visualisation of the example in Figure \ref{fig:sankey_noise_aGMA}, where communities detected by $\alpha$GMA at snapshot 15 remain largely consistent up to snapshot 90, despite minor fluctuations. In contrast, independent GMA suffers from constant changes in community structure and fails to return to the original configuration, as shown in Figure \ref{fig:sankey_noise_GMA}.

sGMA initialisation improves the stability by 7\% for intermittent transformation but leads to a 7\% loss in correctness compared to independent GMA. As illustrated in the example of Figure \ref{fig:sankey_noise_sGMA}, the intermittent noise leads sGMA to merge communities at snapshot 60. Subsequent initialisation maintains these merged communities. Without a refinement, this causes a rearrangement of the existing communities preventing the detection of more communities in the next snapshot and ultimately the reconstruction of the original ones.

Conversely, NeGMA introduces a local modularity evaluation during initialisation. This overcomes the sGMA limitation and offers the best trade-off between correctness and stability. This results in a correctness decrease $<1\%$ and a stability increase of $\approx 3\%$.  Supporting this, \cref{fig:sankey_noise_neGMA} illustrates partial reconstruction of original communities between snapshots 45 to 75.

\paragraph*{Morphing transformations} 

\begin{figure}[!t]
    \centering
    \begin{subfigure}[b]{.5\linewidth}
        \centering
        \includegraphics[width=\linewidth]{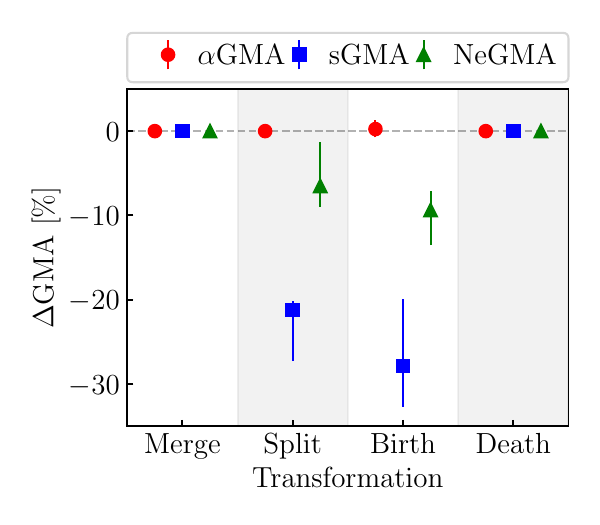}
        \caption{Correctness}
        \label{fig:morphing_correctness}
    \end{subfigure}
    \hspace{-.5em}
    \begin{subfigure}[b]{.5\linewidth}
        \centering
        \includegraphics[width=\linewidth]{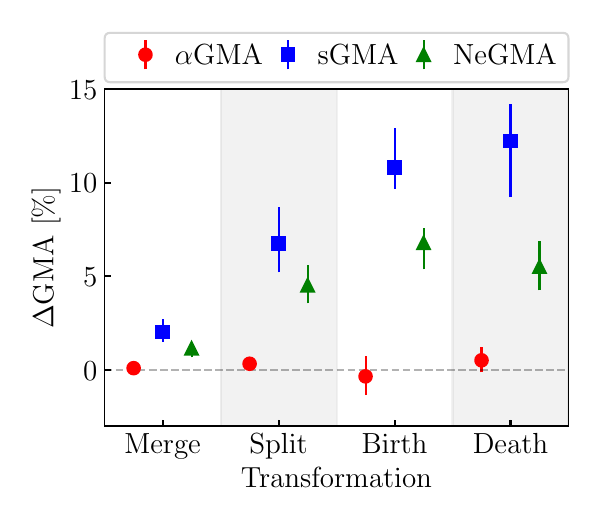}
        \caption{Stability}
        \label{fig:morphing_stability}
    \end{subfigure}
    \caption{\emph{Morphing}. Median gain and bootstrapped 99\% confidence interval in Correctness and Stability.}
    \label{fig:morphing}
\end{figure}

\cref{fig:morphing} reports the \emph{median} correctness and stability gain/loss resulting from morphing transformations. $\alpha$GMA performs comparably to independent GMA across all designed transformations.

Notably, sGMA emerges as the top performer in terms of stability. It achieves a stability improvement ranging from +3\% for merges, up to +13\% for deaths. However, this improvement is countered by a significant loss of $>20\%$ in correctness for split and birth transformations. This limitation is consistent with sGMA challenge in identifying emerging communities.

NeGMA once again proves to strike the best balance between correctness and stability. It achieves an overall maximum stability improvement of 7\% while constraining the losses in correctness to no more than $\approx -10\%$ for both split and birth transformations.

\emph{Takeaway: (i) The memory term of $\alpha$GMA does not bring substantial changes compared to independent GMA, a part from intermittence transformation; (ii) sGMA achieves high stability over time in all the scenarios, but fails in detecting emerging communities; (iii) NeGMA emerges as the most balanced algorithm in morphing transformations, achieving competitive performance in both correctness and stability.}

\section{Responsiveness Evaluation}
\label{ss:detection}

We assess the responsiveness of evolutionary GMA solutions in detecting changes through a new experiment involving morphing transformations. We generate 100 independent graphs for each transformation and run 10 instances of each algorithm for $N=20$ simulated snapshots. We perform an instantaneous (\ie $\tau=1$) transformation at snapshot $t=10$.
We measure the delay $D$ to detect this transformation. Given the setup, $D$ falls in the $[0, 10]$ range and longer delays imply lower responsiveness.

\cref{tab:Delaytable} reports the median delay for each transformation and the fraction of detected transformations (CP is defined). As expected, independent GMA detects the transformation at the exact snapshot with $D=0$ for all the cases. 

The memory term of $\alpha$GMA, in turn, forces the overall graph to evolve at a slower rate, resulting in a systematic delay and lower responsiveness (from $D=1$ for splits to $D=4.6$ for births).

\begin{table}[!t]
\centering
\footnotesize
\setlength{\tabcolsep}{2pt}
\caption{Median delay and reached crossing points in \emph{morphing} scenario. Best results are highlighted in \textbf{bold}, critical results in \color{red}{\textbf{red}}.}
\label{tab:Delaytable}
\begin{tabular}{lcccc|cccc}
\toprule
 & \multicolumn{4}{c|}{\textbf{Delay} $D$} & \multicolumn{4}{c}{\textbf{Reached CP {[}\%{]}}} \\
 & \textbf{Merge} & \textbf{Split} & \textbf{Birth} & \textbf{Death} & \textbf{Merge} & \textbf{Split} & \textbf{Birth} & \textbf{Death} \\
\cmidrule(lr){2-9}
GMA & \textbf{0.0} & \textbf{0.0} & \textbf{0.0} & \textbf{0.0} & \textbf{0.98} & 0.88 & 0.90 & \textbf{1.00} \\
$\alpha$GMA & 3.0 & 1.0 & 4.6 & 2.0 & \textbf{0.98} & 0.90 & 0.90 & 0.96 \\
sGMA & \textbf{0.0} & \color{red}{\textbf{Max}}  & \textbf{0.0} & \textbf{0.0} & \textbf{0.98} & \color{red}{\textbf{0.04}} & \textbf{0.98} & 0.82 \\
NeGMA & \textbf{0.0} & \textbf{0.0} & \textbf{0.0} & \textbf{0.0} & \textbf{0.98} & \textbf{0.94} & 0.88 & \textbf{1.00} \\
\bottomrule
\end{tabular}
\end{table}

\begin{figure}[!t]
    \centering
    \begin{subfigure}[b]{.49\linewidth}
        \centering
        \includegraphics[width=\linewidth]{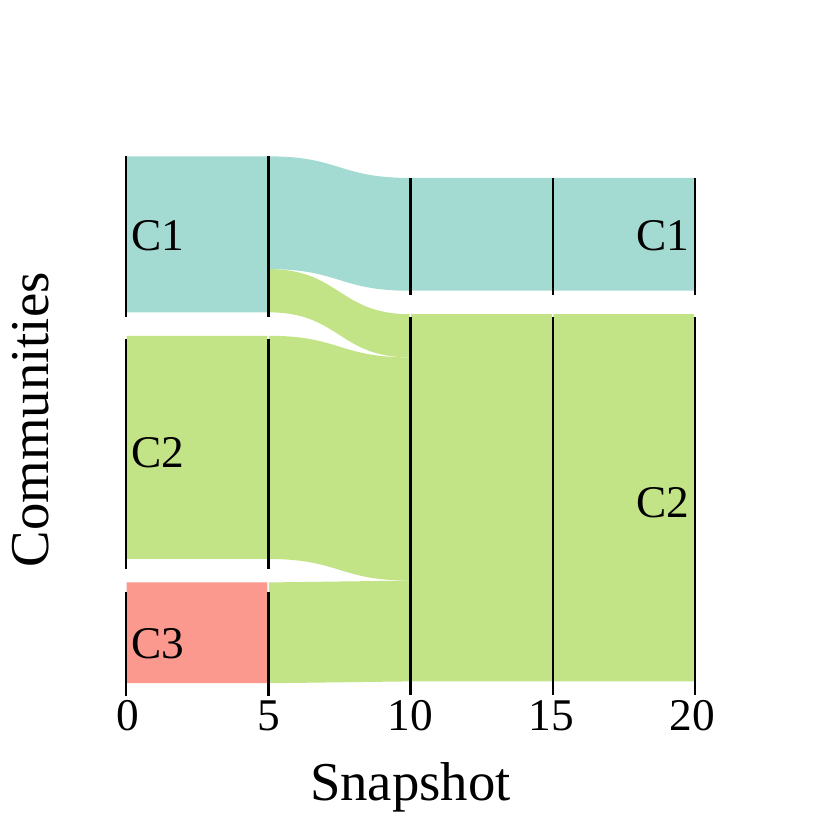}
        \caption{sGMA}
        \label{fig:sankey_sGMA}
    \end{subfigure}
    \begin{subfigure}[b]{.49\linewidth}
        \centering
        \includegraphics[width=\linewidth]{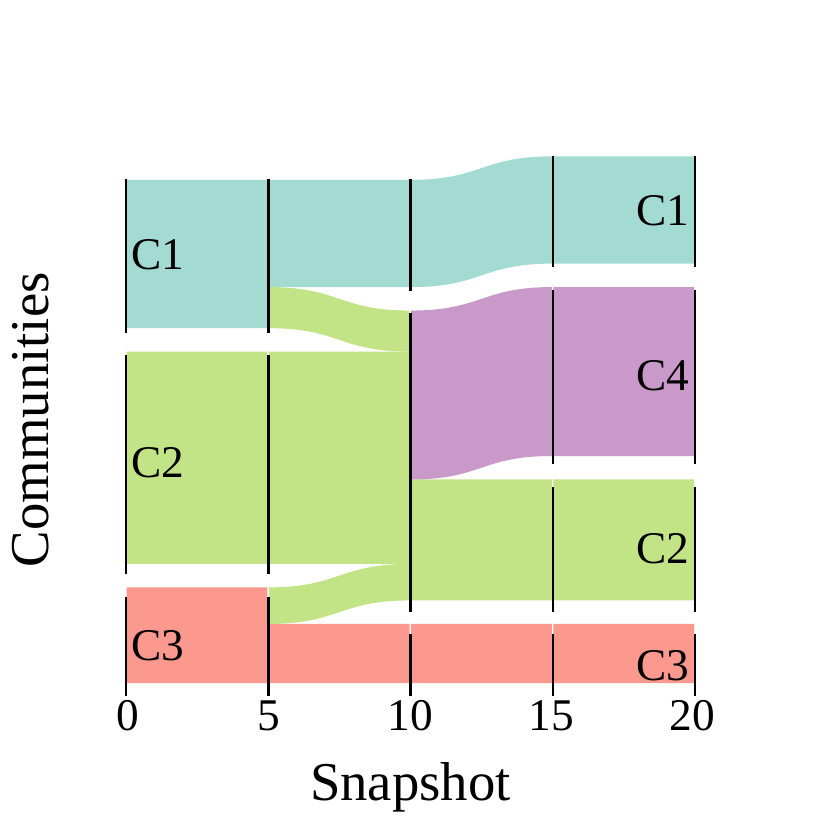}
        \caption{NeGMA}
        \label{fig:sankey_neGMA}
    \end{subfigure}
    \caption{Dynamic evolution of the graph with instantaneous birth transformation at time 10, showing the community assignment. Zoom in a sample of snapshots from 0 to 20. }
    \label{fig:sankey}
\end{figure}

While sGMA matches the delay of independent GMA for merges, births and deaths, it manages to reach only 4\% of the CPs, failing to detect the vast majority (96\%) of the splits (thus $D$ is set to Max) due to the initialisation at each snapshot. Preserving the previous snapshot assignment, sGMA fails to detect an increasing number of communities compared to the initialisation.
It is worth noting that this limitation emerges also for birth transformations. Even though sGMA reaches 98\% of CPs, it remains tied to the ones of GT$^0$ (as highlighted in \cref{fig:sankey_sGMA}) and it fails to detect emerging communities. This results in a correctness variation of $-37.8\%$ on GT$^N$.

Finally, NeGMA detects instantaneous transitions with a median delay $D=0$ for all the cases. Its evaluation of the local modularity $\Delta Q^t_c$ during node initialisation addresses the sGMA limitation, refining the initial assignment and successfully detecting 94\% of splits (best performer for this transformation). Notably, even though NeGMA reaches 10\% fewer CPs than sGMA, it successfully detects the transformations (unlike sGMA), as highlighted in \cref{fig:sankey_neGMA}, resulting in a correctness of 0.97, compared to the 0.60 of sGMA on GT$^N$.

\emph{Takeaway: (i) The memory term of $\alpha$GMA introduces a systematic delay in detecting instantaneous transformations; (ii) sGMA is insensitive to the abrupt formation of new communities; (iii) NeGMA exhibits high responsiveness and detection rates in most of the instantaneous morphing transformations.}

\section{Scalability Evaluation}
\label{ss:runtime}

Finally, we evaluate the algorithm execution times as the graph evolves gradually. $\alpha$GMA exhibits convergence times $\approx 20\%$ slower than independent GMA. Preserving the past connections among nodes through $\alpha$ causes the algorithm to process a growing graph (more nodes and more edges) at each snapshot. This impacts the overall scalability making the $\alpha$GMA less suitable for constantly evolving dynamic graphs.

Conversely, sGMA emerges as the most scalable algorithm compared to the baseline, with an execution time twice as fast. By retaining membership of the previously detected communities for existing nodes, it efficiently bootstraps the algorithm, yielding a sub-optimal solution as initialisation. 

Lastly, NeGMA achieves a convergence time comparable to the baseline. 
Exploring different modularity thresholds or employing a different community assignment of new nodes among their neighbourhood could further enhance NeGMA performance.

\section{Conclusions}
\label{s:conclusions}

In this paper, we addressed the challenge of evaluating evolutionary CD algorithms in dynamic networks. We proposed a benchmarking framework relying on a set of graph transformations which reflect real-world scenarios. We supplied the lack of a comprehensive set of tools to assess the quality of detected communities proposing three new metrics (both supervised and unsupervised) complementary to the widely used modularity. We also proposed NeGMA, a generalised modularity-based evolutionary CD approach relying on GMA and node neighbourhood.
We adopted the proposed framework to extensively test and compare the performance of different evolutionary CD algorithms. 

In a nutshell, experimental results reveal that (i) all the algorithms exhibit robustness against disruptive and noisy scenarios; (ii) while $\alpha$GMA demonstrates both high stability and correctness in detecting intermittent transformations, it introduces high delays in detecting instantaneous transformations and strongly impacts on convergence time; (iii) sGMA excels in scalability and maintains high stability over time (especially in gradual transformations affecting the community structure of the network), but exhibits low responsiveness and struggles in detecting emerging communities; (iv) NeGMA is the most balanced algorithm in terms of correctness and stability. Furthermore, it outperforms the responsiveness and detection rates of existing evolutionary solutions when abrupt changes occur. 

Future developments include extending the designed transformations and testing different combinations of them including also real datasets. A promising extension of the proposed evaluation comprises other modularity-based CD approaches and their application to real-world problems. Additionally, a deeper investigation of the impacts of the parameters involved in the tested solutions can provide further insights into their performance, especially the modularity threshold $\theta_Q$ of NeGMA which could have a strong impact on some transformations, such as expansion.

\section*{Acknowledgement}
This work has been funded by the project "National Center for HPC, Big Data and Quantum Computing", CN00000013 (Bando M42C – Investimento 1.4 – Avviso Centri Nazionali” – D.D. n. 3138 of 16.12.2021, funded with MUR Decree n. 1031 of 17.06.2022).

\bibliography{bibliography}

\end{document}